# A Surrogate Model for Quay Crane Scheduling Problem

Kikun Park[a], Hyerim Bae[b, *]

[a]Safe & Clean Supply Chain (SCSC) Research Center, Pusan National University; kikunpark@pusan.ac.kr

[b]Department of Data Science, Graduate School of Data Science, Pusan National University; hrbae@pusan.ac.kr

**Abstract**

In ports, a variety of tasks are carried out, and scheduling these tasks is crucial due to its significant impact on productivity, making the generation of precise plans essential. This study proposes a method to solve the Quay Crane Scheduling Problem (QCSP), a representative task scheduling problem in ports known to be NP-Hard, more quickly and accurately. First, the study suggests a method to create more accurate work plans for Quay Cranes (QCs) by learning from actual port data to accurately predict the working speed of QCs. Next, a Surrogate Model is proposed by combining a Machine Learning (ML) model with a Genetic Algorithm (GA), which is widely used to solve complex optimization problems, enabling faster and more precise exploration of solutions. Unlike methods that use fixed-dimensional chromosome encoding, the proposed methodology can provide solutions for encodings of various dimensions. To validate the performance of the newly proposed methodology, comparative experiments were conducted, demonstrating faster search speeds and improved fitness scores. The method proposed in this study can be applied not only to QCSP but also to various NP-Hard problems, and it opens up possibilities for the further development of advanced search algorithms by combining heuristic algorithms with ML models.

## 1. Introduction

Container cargo, responsible for 90% of global trade, is transported through ports [19]. In ports, Berth planning is conducted before container ships dock. Berth planning is defined as the Berth Allocation Problem (BAP), which determines the operational time of ships within the limited berth space [27]. At this stage, the number of Quay Cranes (QCs) to be deployed is decided by considering the workload and possible docking times of the incoming ships, which is defined as the Quay Crane Allocation Problem (QCAP). Since QCAP requires creating a plan for ships already docked, it is closely related to BAP [28]. Finally, the issue of determining the start time of all tasks assigned to QCs is known as the Quay Crane Scheduling Problem (QCSP). A crucial characteristic of QCSP is that there must be no physical interference between two or more QCs during operations.

Previous studies have identified that ports face issues with maintaining punctuality, and to address this, it is essential to generate accurate port operation plans [30]. As QC allocation and task planning are closely related to the actual operational time of ships, accurate planning for QCs is crucial in producing a berth schedule that aligns with the actual working time [29].

Container ships entering ports for loading and unloading operations are loaded sequentially according to the unloading and loading requirements, container size, weight, and type in each bay, which is referred to as the Bay Plan [1]. QCSP involves arranging the sequence of crane operations for each ship's bay so that the scheduled operations do not exceed the allocated docking time. In some cases, planning is done with consideration for environmental friendliness, operational costs, and safety, reflecting the operational characteristics of the port [20, 21, 22, 23].

The Quay Crane Scheduling Problem (QCSP) is an NP-Hard problem, requiring complex mathematical formulations and substantial computational time to find an optimal solution while considering various constraints [24]. The fundamental constraints that need to be considered in QCSP are as follows:

· **Constraints on QC working conditions**

This refers to the limitations of the QC's operational environment. For instance, QCs are installed on rails and move left and right within their allowable range. A key constraint is that one QC cannot pass over another. Additionally, it is necessary to prevent situations where QCs' operational areas overlap, or collisions occur during operations.

· **Constraints on the ship's working condition**

Ensuring stability during operations is crucial to prevent the ship's center from shifting. Constraints related to the minimum and maximum number of QCs assigned, based on the ship's specifications, must also be set. In some cases, the berth length and QC's operational range are also taken into account.

· **Assumptions regarding QC operational speed**

QC operational speed is vital for the success of ship operations, and thus predicting the estimated completion time is essential. Previous studies have often assumed a constant speed for QC operations when generating plans. However, in reality, QC operations do not proceed at a constant speed due to various environmental factors (e.g., task type, QC movement, operator changes, yard truck arrival delays). When QC operations are not completed on time, plan adjustments become necessary, and bottlenecks occur. Therefore, accurately predicting QC operational speed is crucial.

· **Consideration of computational complexity**

Given the large computational effort required to solve QCSP while adhering to multiple constraints, generating an optimal plan can sometimes take a long time. Thus, reducing computational complexity is a critical issue in solving this problem.

This paper introduces a method for generating an optimal Quay Crane Scheduling Problem (QCSP) solution given a bay plan. The contributions of this research are as follows. First, unlike previous studies that predetermined the number of Quay Cranes (QCs) to be deployed [9, 10, 16], this study introduces a search method that simultaneously considers both the number of QCs and the scheduling plan. In particular, instead of solving QCSP with a fixed number of QCs, bays, and workload, this study conducts experiments on generating work plans for various bay plans. For these experiments, a novel two-dimensional chromosome representation is introduced.

Second, the study proposes a method to generate more precise work plans by learning from real-world port data to more accurately calculate QC operational time. It was found that using a Machine Learning (ML) model-based prediction approach yields more accurate plan generation than relying on fixed constants or probabilistic simulations.

Lastly, a Surrogate Model which combines a GA from evolutionary algorithms with an ML model, is introduced to generate the optimal QC plan. The Surrogate Model technique accelerates search speed by training the fitness function, and it has been demonstrated to offer faster computation times than previously used GA methods for QCSP.

The remainder of this paper is structured as follows: Chapter 2 reviews the related studies on QCSP and GA methods used to solve NP-Hard problems. Chapter 3 introduces the objective function and constraints of the QCSP as defined in this study, along with the ML-based method for predicting QC operational speed. Chapter 4 discusses the Surrogate Model approach for generating the optimal QC plan. Chapter 5 summarizes the experimental results, comparing the accuracy of QC operational speed predictions with and without the use of ML, as well as the application of the Surrogate Model and previous GA methods to QCSP. Finally, the conclusion is discussed.

## 2. Related works
### 2.1. Considerations for the Quay Crane Scheduling Problem (QCSP)

QCSP (Quay Crane Scheduling Problem) is a critical issue directly related to the operational tasks of ships, and extensive research has been conducted on this topic. Fundamentally, it is known that QCSP involves considering four key attributes. The first is the **task attribute**, which defines the workload of the vessel, including bay areas and single container movements. The second is the **crane attribute**, which refers to the specifications of the crane, such as its preparation time and initial position, as well as whether the crane's movement speed is taken into account. The third attribute is **inference**, which encompasses the movement constraints and safe distance requirements of the quay cranes. Lastly, the **performance measure** attribute pertains to the target objectives of QCSP, such as minimizing the completion time of operations [3].

Initially, research focused on the movement and constraints of quay cranes [4]. Subsequent studies incorporated additional considerations, such as the vessel's berthing and departure times, non-crossing constraints for cranes, and the need to maintain safe distances [5]. As ship sizes and workloads continued to increase over time, the complexity of the cases to be considered in QCSP grew exponentially, leading to the development of new approaches to address these challenges [6]. Additionally, as the number of containers to be handled by the cranes increased, a maintenance period model for quay cranes, which emphasized the importance of maintenance scheduling, was introduced [7].

Furthermore, the potential imbalance in the ship's load during container discharging operations, which could cause the vessel to tilt either forward or aft, became a critical constraint to consider when planning the discharging process [8]. Research later extended to planning that involved not only discharging but also simultaneous loading, while accounting for draft-trim constraints [9, 10]. While current quay cranes can only work on a single bay at a time, a QCSP model that anticipates future quay cranes capable of operating on two bays simultaneously has been introduced [14]. In addition, studies have expanded performance measures to include not only crane operating speed but also factors such as driver costs, pollution emissions, and energy consumption [20-23].

When creating operational plans at ports, considering not only QC operations but also those of yard trucks (YT), yard cranes (YC), and yard blocks can yield solutions closer to real-world conditions. In this context, an approach that simultaneously considers QC schedules and yard block loading conditions has been proposed [11]. Similarly, research has been conducted that integrates the scheduling of YTs responsible for container transport within the yard [12]. Moreover, there are studies that address QCSP in conjunction with the Berth Allocation Problem [13, 25].

## 2.2. Assumptions for Quay Crane Work Speed

Given the central role of QC in ship operations, assumptions regarding their operating speed are of critical importance. Many prior studies on the QCSP have assumed a constant speed for QCs [7, 8, 15, 16]. To address the need for a more accurate estimation of QC cycle time, a statistical model has been introduced [17]. As mentioned earlier, during real-world ship operations, various factors can cause delays. In response to this, research has been conducted that incorporates uncertainties in the operating speeds of both QCs and straddle carriers (SC) to solve the QCSP more effectively [18].

In order to predict ship operation times based on a given work plan, researchers have emphasized the use of machine learning (ML) models leveraging historical operational data. These studies suggest that properly utilizing ML predictive models can significantly reduce prediction errors [2, 31]. In this study, rather than assuming a constant or random variable for QC speed, a more accurate QC scheduling plan was generated by utilizing an ML model.

### 2.3. Computational Complexity of Quay Crane Scheduling Problem

It is widely known that finding an optimal solution for the QCSP requires extensive computational effort, and various efficient search methods have been proposed to address this challenge. Initially, the Branch and Bound (B&B) method was introduced to efficiently solve QCSP by systematically selecting feasible child nodes [4]. Later, researchers demonstrated that QCSP is an NP-Hard problem, which led to discussions on computational complexity not covered in previous work [4]. In response, studies began employing Genetic Algorithms (GA) to tackle the problem [24]. GA has been extensively utilized in QCSP, and several methods have been proposed to improve its performance.

To reduce the search time in QCSP, a modified GA with roulette wheel selection was introduced [6]. Additionally, a customized GA was developed to solve both the Quay Crane Assignment Problem (QCAP) and QCSP simultaneously. This approach reduced computation time by adding feasible individuals to the randomly generated initial population [26]. A simulation-based GA, combining Monte Carlo simulations with GA, was also proposed [18].

When applying the generated QC work plans to real-world terminal operations, the constraints need to be realistic, which significantly increases the computational complexity for large problems. Therefore, it has been argued that methods capable of efficiently handling such large problems are necessary [8]. To address this, researchers introduced a Hybrid GA that reduces the search space and computation complexity, particularly for large problems, which require different search methods compared to small and medium-sized problems suitable for B&B methods [10]. Similarly, studies have been conducted on reducing the feasible solution space using bicriteria evolutionary methods [9].

Later, a mathematical model utilizing Logic-Based Benders Decomposition was introduced to solve QCSP by breaking down the larger optimization problem into smaller sub-problems. This model reduced computational cost by checking the feasibility of constraints (except for vessel stability) first, and only verifying the stability constraint for schedules deemed feasible [16]. However, traditional GA methods used for QCSP have a drawback in that they require evaluating a large number of chromosomes. To address this issue, this study introduces a surrogate model that reduces the evaluation time by learning the GA fitness function in advance, thereby narrowing the search space.

## 2.4. Surrogate Model for Quay Crane Search Problem

As discussed in Section 2.3, GA are widely used to solve NP-Hard problems such as the QCSP. GA involves several key processes, including population generation, crossover, and mutation, during which the fitness score of chromosomes must be calculated. The size of the generated chromosomes significantly influences the computational effort required [32]. In particular, NP-Hard problems like QCSP, which involve creating time windows for scheduling and determining the feasibility of a given plan, can require substantial time just to calculate the fitness score for a single chromosome. Since finding the optimal solution necessitates evaluating a large number of chromosomes, reducing computational costs is essential.

In this study, we propose a method to reduce the search time of GA by pre-training the fitness function. The concept of training the fitness function has been employed in other fields as well. For example, in research on solving the Intensity-Modulated Radiation Therapy (IMRT) beam angle optimization problem using GA, the traditional chromosome evaluation method was found to be time-consuming. To address this, a neural network was trained to predict the fitness function, thereby reducing the time spent on chromosome evaluation and shortening the overall search process [33]. Similarly, in a study on solving molecular design problems using GA, a Deep Neural Network (DNN) was utilized to assess the fitness score, while a Recurrent Neural Network (RNN) Decoder was used to redesign molecules, demonstrating the advantage of being able to evaluate molecular suitability even under complex constraints [34].

In real-world engineering environments, more heavily constrained problems have been tackled efficiently by combining Rule-Based Reinforcement Learning (RL) with GA, leading to improved computational efficiency [35]. In this research, we propose a method for training the GA fitness function in the context of QCSP, which aims to reduce the search space and find the global optimum more efficiently.

## 3. Problem Description and Mathematical Formulation

In this chapter, we will address the mathematical formulation required to define the operational approach and schedule generation for QCs.

### 3.1. Quay Crane Scheduling

According to Figure 1, each hatch of the vessel can accommodate either two 20ft containers or one 40ft container, with the QC primarily positioned at the hatch to carry out its operations. The QC moves between hatches to perform loading and unloading operations, and there is a constraint that prevents multiple QCs from operating on the same hatch at the same time. Therefore, the scheduling of QC operations is planned based on the allocation of tasks to specific hatches.

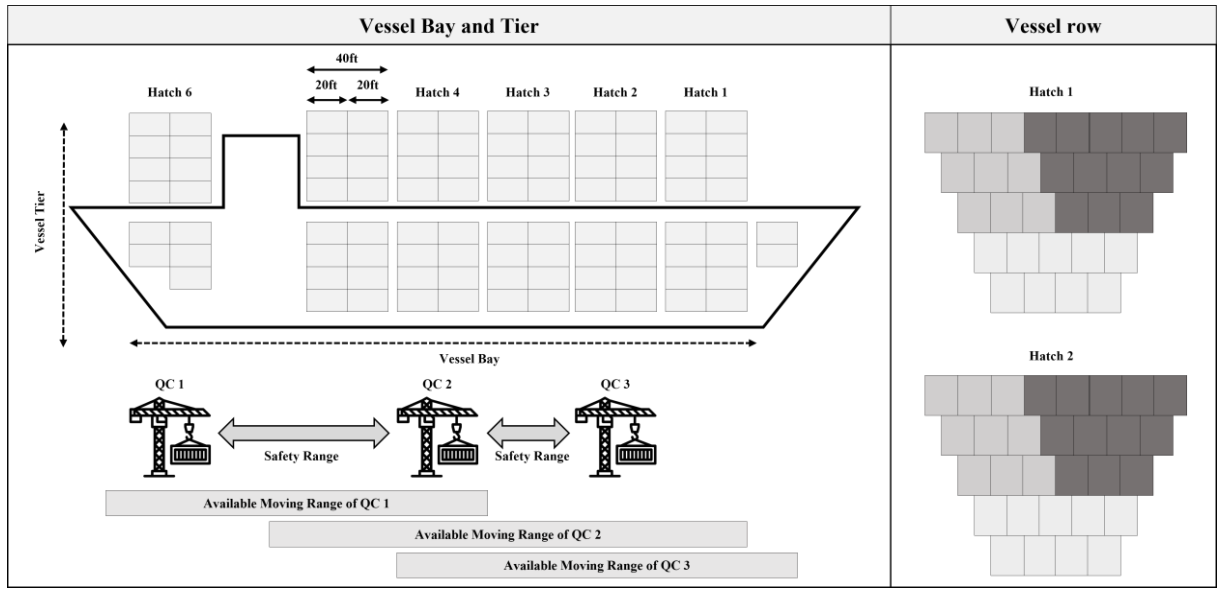

**Figure 1. Layout of container vessel**

The QC schedule generated in Figure 2 is defined as the bay plan, and the bay plan consists of multiple tasks, as illustrated in Figure 2. Each task represents a specific operation assigned to a particular bay, contributing to the overall execution of the plan.

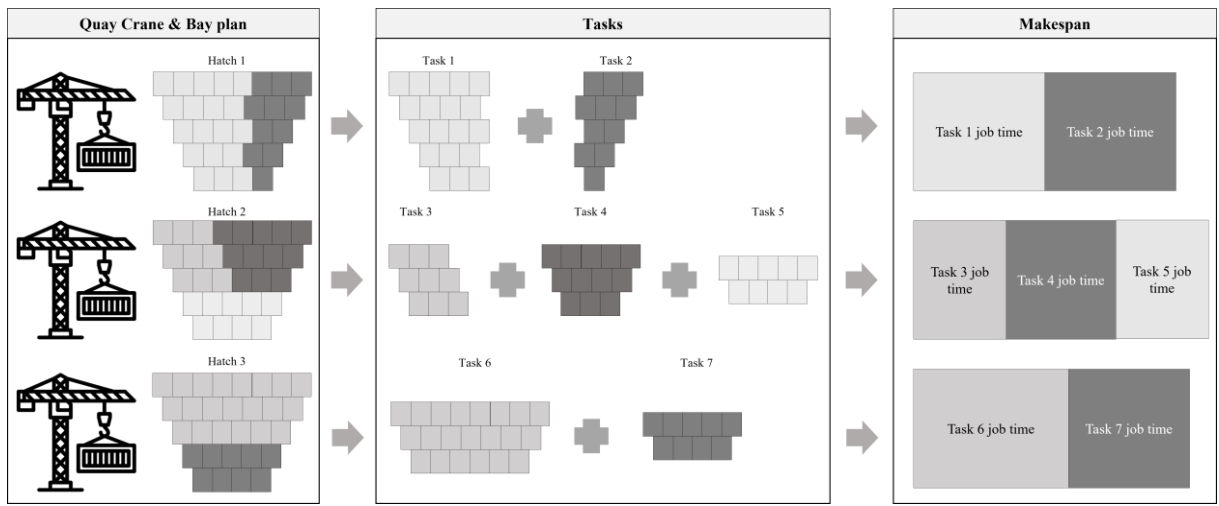

**Figure 2. Bay plan and tasks**

In this study, we conducted experiments to assign the sequence of operations for quay cranes (QC) based on the pre-planned bay plan and tasks. The primary factor to consider when determining the sequence of QC operations is the makespan of the assigned QCs. Makespan refers to the total time elapsed from the start of the first task to the completion of the final task for a given vessel, and it is crucial that the makespan does not exceed the vessel's scheduled berthing time. While assigning more QCs can expedite the completion of operations, it also increases operational costs. Therefore, in this research, we established an objective function aimed at minimizing the difference between the berthing time of the vessel and the makespan, ideally bringing it as close to zero as possible.

In generating the optimal QC operation sequence, we considered various constraints that arise in practical operational environments. First, since QCs are installed on rails at the berth, they can only move horizontally (left and right), and each QC is limited to specific operational zones. In this study, we set the movement direction of QCs as left to right, and for each individual QC, we defined zones where it can and cannot operate based on the location of the vessel bays. Additionally, due to spatial constraints, no more than one QC can operate on a single bay at the same time, and QCs are prohibited from crossing over each other during movement. Lastly, the time required for a QC to complete a task was predicted using a ML model, as applied in previous studies [2]. The formulation of this model is expressed mathematically as follows.

**Indices:**

$t$: Index of time during vessel handling time

$b$: Index of vessel bay number, arranging in an increasing number from stem to stem

$j$: Index of tasks which are increasing order according to bay plans

$i$: Work sequence of task $j$

$k$: Index of quay crane number which is an increasing order according to their relative locations (left to right)

**Decision Variables:**

$H$: Nonnegative continuous variable representing the expected time of berthing time

$C$: Nonnegative continuous variable representing the makespan

$s_j^k$: Nonnegative continuous variable representing the start time of task $j$

$s_j$: Nonnegative continuous variable representing the minimum value of $s_j^k$

$c_j^k$: Nonnegative continuous variable representing the completion time of task $j$

$c_j$: Nonnegative continuous variable representing the maximum value of $c_j^k$

$\alpha$: A parameter for setting the weight of $C$, between 0 and 1

$n_t^b$: Number of QCs in hatch $b$ at time $t$

**Input parameters:**

$ew_j^k$: Estimated time of work for QC $k$ to process task $j$

$ec_{ji}^k$: Estimated time of work for QC $k$ to process transfer $i$-th sequence container of task $j$

$con_{ji}^k$: $i$-th sequence container in task $j$ assigned to QC $k$

$\delta_{ji}^k$: Unloading/Loading of $con_{ji}^k$, if loading 1, else 0)

$tw_{ji}^k$: Single/Twin of $con_{ji}^k$, if twin 1, else 0

$fc_{ji}^k$: Full/Empty of $con_{ji}^k$, if full 1, else 0

$ct_{ji}^k$: Container type of $con_{ji}^k$, if Danger, Bundle and Reefer 1, else 0

$jc_{ji}^k$: Change job of $con_{ji}^k$, if change 1, else 0

$q_t^k$: Location of QC $k$ at time $t$

$l_k$: The leftmost bay of QC $k$ can move

$r_k$: The rightmost bay of QC $k$ can move

$b_j$: Container location of task $j$

**Formulations:**

$$\min |C - H| \tag{1}$$

$$C = c - s \tag{2}$$

$$s_j^k = \begin{cases} s_1^k, & \text{if } j = 1 \\ c_{j-1}^k, & \text{if } j > 1 \end{cases} \tag{3}$$

$$c_j^k = s_j^k + ew_j^k \tag{4}$$

$$ew_j^k = \sum_{i=1}^{I} ec_{ji}^k \tag{5}$$

$$ec_{ji}^k = f_1(\delta_{ji}^k, tw_{ji}^k, fc_{ji}^k, ct_{ji}^k, jc_{ji}^k) \tag{6}$$

$$q_t^1 < q_t^2 < \cdots < q_t^k, \quad k > 2 \tag{7}$$

$$n_t^b < 2 \tag{8}$$

$$l_k \leq q_t^t \leq r_k \tag{9}$$

Equation (1) represents the target objective function, where $H$ is the scheduled berthing time of the vessel and $C$ is the makespan for the vessel, with the objective being to minimize the absolute difference between these two values. In Equation (2), $C$ is calculated as the time elapsed from the start of the first QC operation to the completion of the final task in the bay plan. The completion time ($c$) is determined as the maximum completion time ($c_j^k$) for QC ($k$) performing task ($j$). The completion time ($c_j^k$) is calculated by adding the task's processing time ($ew_j^k$) to the start time ($s_j^k$), as summarized in Equations (3) and (4). For the first task of QC ($k$) (i.e., $j = 1$), the start time ($s_1^k$) is defined as the initial time of QC's deployment. Otherwise, the start time for subsequent tasks is defined by the completion time of the previous task.

The processing time ($ew_j^k$) for QC ($k$) to complete task ($j$) can be expressed as the sum of the processing times for individual containers ($ec_{ji}^k$). Thus, Equations (5) and (6) detail the process of predicting the total processing time ($ew_j^k$) for task ($j$) and the processing time ($ec_{ji}^k$) for the $i$-th container ($con_{ji}^k$) in task ($j$). To predict the processing time ($ec_{ji}^k$) for each container, an ML model was employed, denoted as $f_1$. The ML model $f_1$ can utilize various algorithms suitable for predicting numerical values, such as linear regression (LR), random forest (RF), support vector regressor (SVR), or multiple layer perceptron (MLP). The features used to predict the processing time for a container ($ec_{ji}^k$), as introduced in previous research [2], include five key factors in this study:

- $\delta_{ji}^k$ indicates whether the container ($con_{ji}^k$) is being loaded or unloaded, where 1 represents loading and 0 represents unloading.
- $tw_{ji}^k$ is defined as 1 if the container is a twin container.
- $fc_{ji}^k$ is defined as 1 if the container is full, and 0 if it is empty.
- $ct_{ji}^k$ represents whether the container is a special type, such as a refrigerated or bundled container.
- $jc_{ji}^k$ indicates whether the task requires a change in operation type (e.g., from loading to unloading).

Equation (6) also introduces $q_t^k$, which represents the position of QC $q_t^k$ at time $t$, ensuring that QCs cannot cross over each other. Equation (8) provides a constraint that ensures no more than two QCs can be present at the same hatch $b$ at time $t$. Finally, Equation (9) defines the range within which QC ($k$) can move. QC ($k$) is restricted to operate within the predefined leftmost boundary ($l_k$) and the rightmost boundary ($r_k$), ensuring that its movement is confined to this region.

## 4. Methodology

In this chapter, we introduce the method of a Surrogate Model that utilizes a learned fitness function to optimize the search process for generating an optimal QCSP solution. By training the fitness function in advance, the Surrogate Model improves the efficiency of the search, reducing the computational effort required to find the optimal solution.

### 4.1. Encoding of Chromosomes for Surrogate Model

In this study, a GA-based Surrogate Model is employed to find the optimal solution for the QCSP. The chromosome representation used in this research is shown in Figure 3. To account for the priority of tasks in each hatch, the study introduces a two-dimensional matrix form of chromosome representation. This approach allows for an efficient encoding of task sequences while considering the operational priorities for each hatch.

In Figure 3, the plot on the left represents the Bay plan, where tasks that need to be performed for each hatch are listed. The plot on the right illustrates the assignment of QC numbers to each task, indicating which QC is responsible for performing the corresponding tasks in the bay plan. When generating a chromosome, the determination of the number of QC units to be deployed and the development of the work plan must be carried out simultaneously. This process can be observed in Algorithm 1.

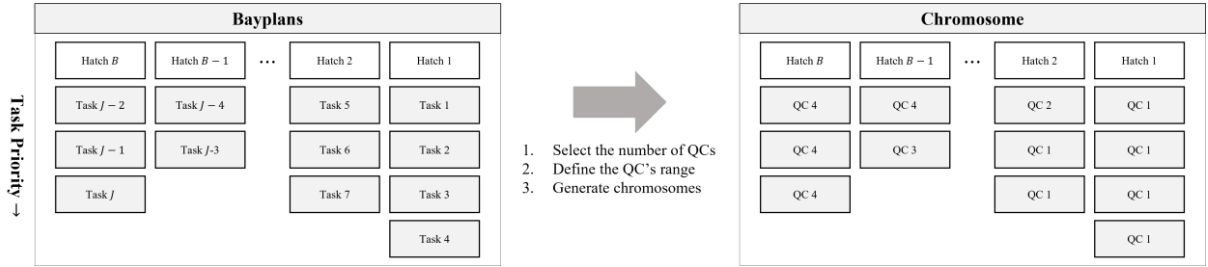

**Figure 3. Encoding scheme**

---

**Algorithm 1 Generating chromosome Set**

**Input**:
    Bay plans of target vessel: List of tasks $1 \sim J$ of each hatch $1 \sim B$
    The number of chromosomes to generate: $nc$
    The minimum number of QC allocated: $lq$
    The maximum number of QC allocated: $uq$

**Output**:

1: $n \leftarrow 0$; Set = [];
2: **while** $n$ equals to $nc$ **do**
2:   Decide the number of QC allocated ($nq$) between $lq$ and $uq$ randomly (using Uniform distribution);
3:   Define the QCs' available movement range: $l_k$ and $r_k$;
4:   Generate a random chromosome ($g$) considering formula (6) ~ (8);
5:   **if** generated chromosome not in Populations **do**
6:     Set[$n$]$\leftarrow g$;
7:     $n \leftarrow n + 1$;
8: **end while**
9: **return** Set

---

According to Algorithm 1, when a bay plan is given, the number of chromosomes to be generated ($nc$) is first determined, followed by the setting of the minimum ($lq$) and maximum ($uq$) number of QC units to be deployed. Next, the range of hatch movement for the QC is established in accordance with the constraints of Equation (9), after which the process of randomly generating chromosomes begins.

The method for evaluating the generated chromosomes is described in Algorithm 2.

---
**Algorithm 2 Evaluating chromosome**

**Input**:
   Chromosome: g
   Vessel berthing time: H
   Features of $con_{ji}^k$: $\delta_{ji}^k, tw_{ji}^k, fc_{ji}^k, ct_{ji}^k, jc_{ji}^k$

**Output**:

1: Predict the processing time of $con_{ji}^k$ (equals to $ec_{ji}^k$) using $f_1$;
2: Calculate the processing time of task $j$ (equals to $ew_j^k$) using formula (5);
3: Get QC tasks start time: $\{s_j^k\}$ and QC tasks completed time: $\{c_j^k\}$ using formula (3), (4);
4: Check the number of QC crossover situations during $s = \min s_j^k$ between $c = \max c_j^k$;
5: Check the two or more QCs are in one hatch at time $t$ (equals to $n_t^b$);
6: **if** the number of QC crossover = 0 and $n_t^b < 2$ **then**
7:   $C = c - s$
7:   **return** $|H - C|$

---

The criteria for evaluating the generated chromosomes are, first, whether a QC crossover situation occurs, and second, whether more than two QCs are positioned at the same hatch simultaneously. If a schedule fails to meet either of these two conditions, it is defined as infeasible. For chromosomes that satisfy both conditions, the absolute difference between the vessel berthing time ($H$) and the makespan ($C$) is calculated as the fitness score. The next chapter will explain the Surrogate Model proposed in this study to improve the search speed of the GA.

### 4.2. Surrogate Model for Quay Crane Scheduling Problem

The Surrogate Model is an algorithm designed to reduce the search space of a GA by training the fitness function. The detailed procedure is outlined in Figure 4, which follows the sequence of the data generation step, fitness function learning step, reducing sample size, and then the genetic algorithm.

In the data generation step, chromosomes are generated according to Algorithm 1 based on the bay plans of ships selected for the training set. Next, the generated chromosomes are evaluated according to Algorithm 2. During the evaluation, feature extraction is performed to store the bay plan information, which is represented as matrices of different dimensions. This process maps the characteristics of chromosomes, expressed in various dimensions, into a data form that can be used for learning. In this study, we propose a method that allows the search to proceed seamlessly even when a new bay plan of a different dimension is input, through feature extraction.

In the fitness function learning step, the chromosomes stored in the first step are trained. Various machine learning (ML) models are applied, and the best-performing algorithm is selected. In the reducing sample size step, chromosomes generated from the bay plans of ships in the test set are not evaluated from the beginning. Instead, the trained ML model is used to select likely candidates. This approach prioritizes the evaluation of chromosomes with a predicted high fitness score, allowing the algorithm to quickly identify chromosomes with high fitness.

Finally, in the genetic algorithm step, the chromosomes selected as likely candidates in the previous step are evaluated, and those deemed feasible are chosen as parents. The GA then proceeds with selection, crossover, and mutation in order to search for the optimal solution to the QCSP.

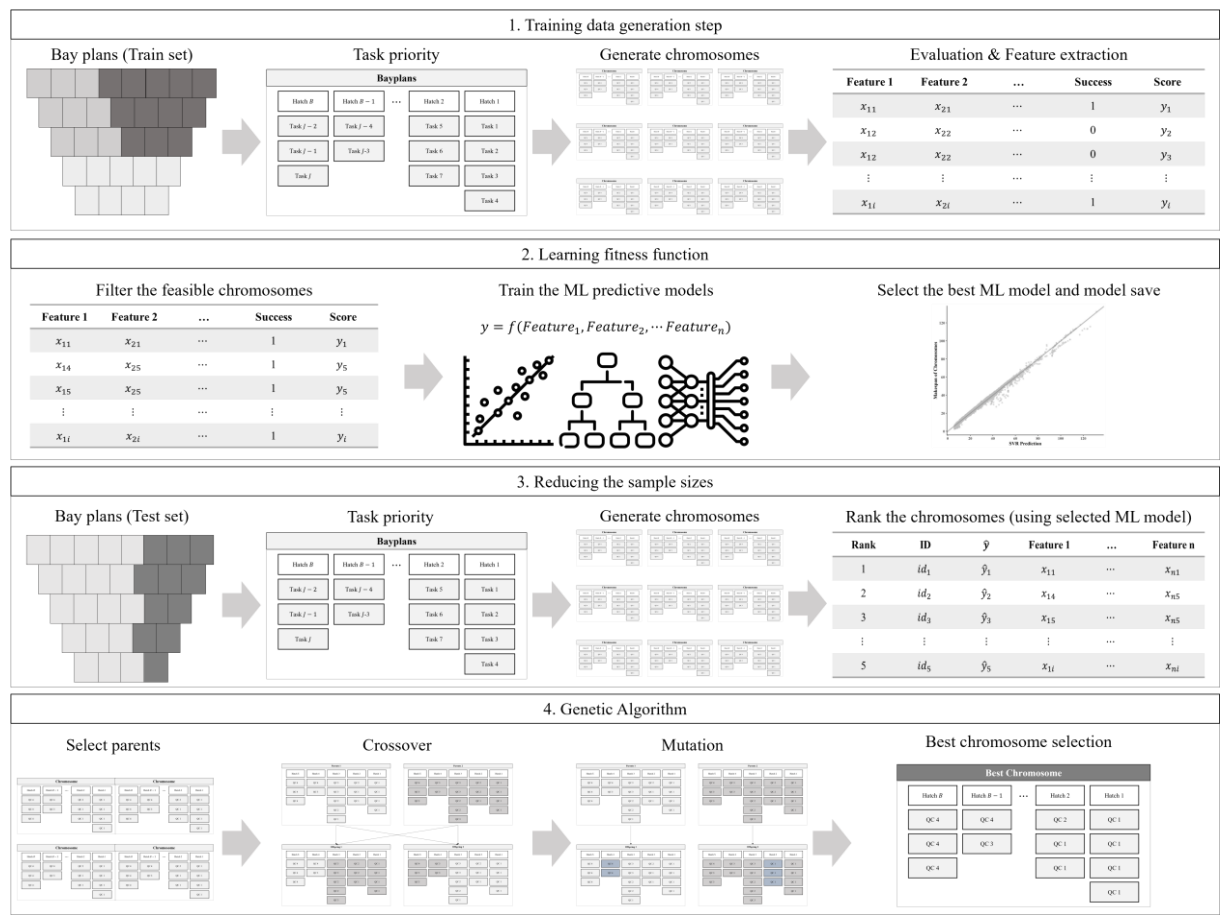

Figure 4. Surrogate Model for QCSP

## 4.3. Training Data Generation

The first step of the Surrogate Model proposed in this study to solve the QCSP is the process of generating training data. To generate the training data, task priorities are defined for the bay plans selected as the test set, and a chromosome set is generated according to Algorithm 1. The generated chromosome sets are then evaluated based on Algorithm 2, and the results are stored as data. This process is repeated.

Table 1. Features for training fitness score

| Notations | Description |
|---|---|
| $CON$ | Total counts of containers |
| $B$ | Number of vessel hatches |
| $p^k$ | Percentage of workload (tasks) allocated per each QC $k$ |
| $q_1^k$ | Initial location of QC $k$ |
| $y_1$ | Feasible or infeasible |
| $y_2$ | Fitness score of chromosomes |

To store the evaluation results as data, the features of the evaluated chromosomes, their feasibility status, and the fitness score are calculated. The features and scores stored in the training data can be found in Table 1.

According to Table 1, $CON$ represents the total number of containers to be processed in the target bay plan, while $B$ refers to the number of hatches on the vessel. The proportion of tasks assigned to a deployed QC ($k$) is denoted as ($p^k$). Additionally, ($q_1^k$) represents the initial working position of QC ($k$). Lastly, ($y_1$) indicates whether the chromosome is feasible, and ($y_2$) represents the fitness score. Once the evaluation and data storage for the chromosomes selected in the training set are complete, the next step involves training the fitness function using a ML algorithm.

### 4.4. Learning fitness function

In the fitness function learning step, the process involves training the fitness function using the train data generated in the previous phase. Only the features of chromosomes deemed feasible are selected, and then the train data is further split into training and testing datasets for the purpose of learning the fitness function. Several ML models are applied to this data, and the prediction results of the fitness scores are compared. The ML model with the best prediction performance is selected and utilized in the Surrogate Model, where it is defined as $f_2$.

### 4.5. Reducing the Sample Size and Genetic Algorithm

Once the fitness function has been successfully learned by the ML model, the Surrogate Model is implemented. The Surrogate Model consists of two main steps: the reducing sample size step and the genetic algorithm step. In the reducing sample size step, task priorities are defined for the bay plan that is being solved for QCSP, and a chromosome set is generated. The pre-trained ML model is then used to predict the fitness scores of the generated chromosome set, which are subsequently ranked. The top-ranked chromosome sets are subjected to actual evaluation to determine their feasibility and calculate their fitness scores. If a chromosome is found to be feasible during this evaluation, it is included in the parents set. Once the pre-determined number of parents has been selected, the process moves to the genetic algorithm step, where crossover and mutation are repeated to generate the best chromosome. The search process of the Surrogate Model is detailed in Algorithm 3.

---

**Algorithm 3 Surrogate Model**

Set maximum generation: $G$
Set number of parents choosing: $np$
Set maximum evolution: $e$
1: $g \leftarrow 0$;
2: **while** $g$ equals to $G$ **do**
3:    Generating chromosome Set();
4:    Predict the fitness scores of chromosomes using ML model ($f_2$);
5:    Rank the chromosomes using the predicted fitness score;
6:    $parents \leftarrow 0$; $genes$=[]
7:    **while** $parents$ equals to $np$ **do**
8:      Evaluate the chromosomes with higher rankings first; # Using Evaluating chromosome()
9:      **If** chromosome is feasible **then**
10:       $genes[np] \leftarrow$ chromosome;
11:       $np \leftarrow np + 1$;
12:      **end while**
11:    **for** $evolution \leftarrow 0$ **to** $e$
11:      Crossover();
14:      Mutation();
15:      Selection();
16:      Update();
17:    $g \leftarrow g + 1$
17: **end while**
18: **return** Best solution;

---

According to Algorithm 3, the parameters for searching for the best chromosome include setting the maximum number of generations ($G$), determining how many parent chromosomes to select ($np$), and defining the number of repetitions for crossover, mutation, and selection ($e$). Once the search begins, the Surrogate Model generates a new initial population according to Algorithm 1 at the start of each new generation.

Since GA are known to have a tendency to get trapped in local minima, the Surrogate Model addresses this by generating new chromosomes with each generation to avoid this issue. However, evaluating a large number of chromosomes would significantly increase the search time. To mitigate this, the pre-trained model ($f_2$) from the training step is used to predict the fitness scores of the newly generated chromosomes. Based on these predicted values, the chromosomes are ranked, and the evaluation of the chromosomes begins with the highest-ranked ones, following Algorithm 2.

The evaluation continues until the number of feasible chromosomes collected reaches the pre-determined number ($np$). Once ($np$) feasible chromosomes are gathered, they are selected as the parents. Crossover, mutation, selection, and update operations are then performed for ($e$) repetitions to search for the best chromosome.

- **Crossover** refers to recombining parent chromosomes.
- **Mutation** introduces random variations (mutations) in chromosomes.
- **Selection** is the process of choosing the chromosome with the highest fitness score as the best chromosome.
- **Update** refers to the function that updates the best chromosome if a better one is found as the generations progress.

If the search reaches the maximum generation ($G$), the process terminates, and the chromosome with the highest fitness score is presented as the solution.

### 4.6. Crossover

Crossover in GA expands the search space by combining existing chromosomes to generate new ones. However, for solving QCSP, it is important to adopt a crossover method that can simultaneously consider both task priority and crane assignment, as noted in previous studies [9, 10]. If crossover is performed randomly in a conventional way, it can lead to issues where the validity of the newly generated offspring cannot be guaranteed. To address this, the Surrogate Model selects two chromosomes randomly from the parents and combines half of each to produce two new offspring. This approach ensures that the newly created chromosomes maintain the necessary balance between task priority and crane assignment.

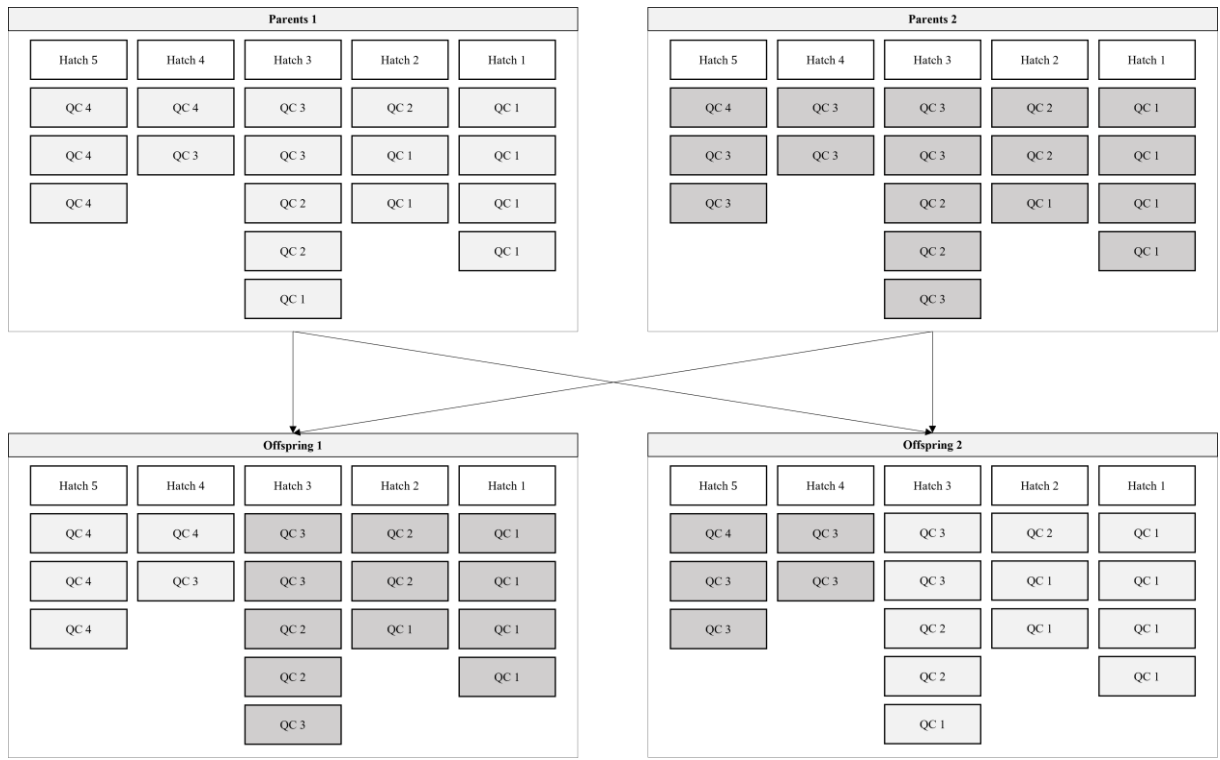

**Figure 5. Crossover**

Figure 5 illustrates an example of crossover, where two randomly selected parent chromosomes are divided at the midpoint of the vessel's hatch and then recombined. The proposed crossover method ensures that the new chromosome generated takes into account both the task priority and the constraints, such as the QC's allowable movement range. This approach enables the creation of a chromosome that respects the key requirements of QCSP while enhancing the search for feasible and optimal solutions.

### 4.7. Mutation

In GA, the purpose of mutation is to provide new solutions in order to avoid getting trapped in local optima, as noted in previous studies [10]. In the Surrogate Model, mutation is performed by randomly selecting a single hatch and reassigning the QC operations for that hatch. This approach introduces variation into the solution space, helping to explore new possibilities while maintaining the structure of the overall chromosome.

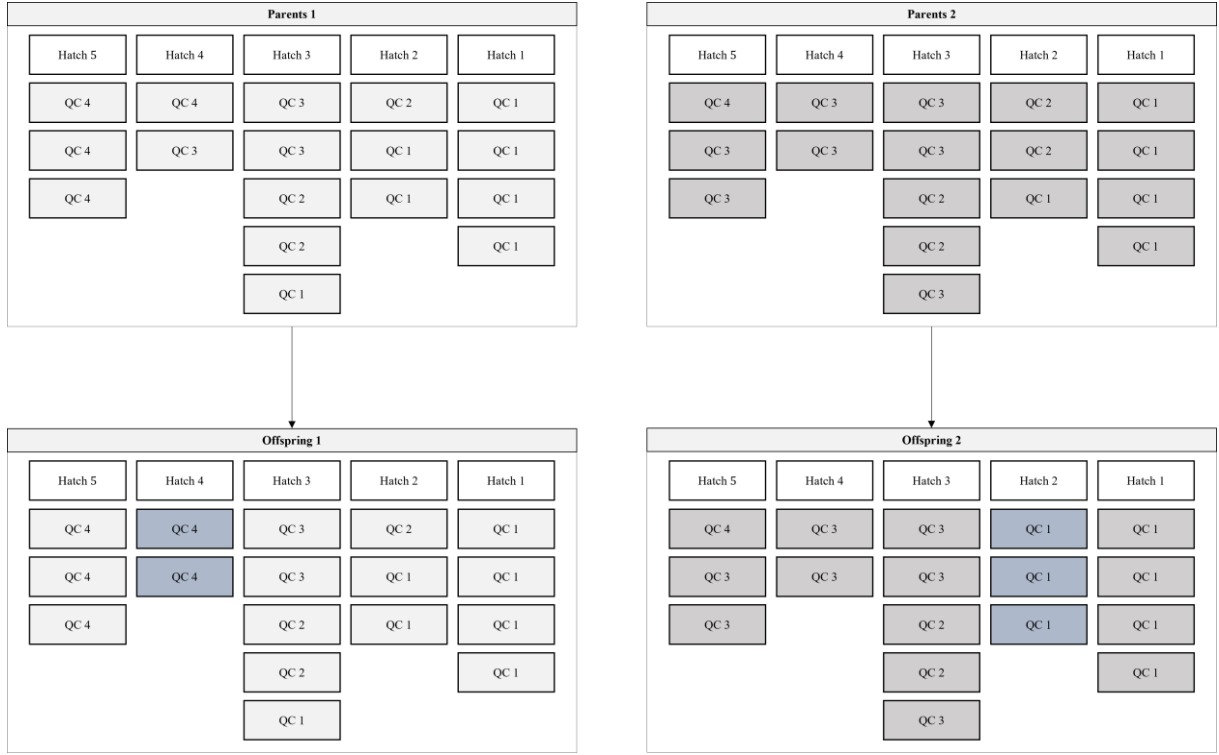

**Figure 6. Mutation**

Figure 6 provides an example of the mutation process, where a random hatch is selected for the two chosen parent chromosomes, and the QC assignment for that hatch is reconfigured. Similar to the crossover strategy, the mutation approach also takes into account constraints such as task priority and the QC's movement range. This ensures that while new solutions are introduced to prevent local optima, they still adhere to the necessary operational constraints of QCSP.

## 5. Experiments

In this chapter, we summarize the experimental results to demonstrate the performance of the proposed method. First, we compare the accuracy of QC work planning when using fixed constants versus when utilizing an ML model trained with container information. Second, we explain which ML model best predicts the fitness function during the training step of the Surrogate Model and introduce the criteria for selecting the prediction model. Third, we conduct a comparative analysis of the Surrogate Model's performance by comparing it with two GA methodologies previously used in studies on QCSP [9, 22]. These earlier studies introduced chromosome representation methods, crossover, mutation, local search procedures, and selection strategies suitable for QCSP.

In this experiment, we performed a comparative analysis between the Surrogate Model and other methods to solve the search problem of determining both the number of QCs and the work plan simultaneously. For the experiment, we used actual port operation data collected from Busan Port

Terminal (BPT) in Busan, South Korea, as training and validation data.

## 5.1. QC Work Speed Prediction

In QCSP and similar NP-hard problems such as the Berth Allocation Problem (BAP) and the Quay Crane Assignment Problem (QCAP), the assumption regarding the QC's working speed is a critical issue. Therefore, this section compares the error in the vessel work plan when using a constant speed versus an ML model to predict the QC's work speed. The experimental results summarize the prediction errors for the time required to handle a single container, the prediction error at the task level, and the prediction error in the makespan for the overall vessel operation.

For the constant settings, the discharge operation was set at 90 seconds per container and the loading operation at 150 seconds. Additionally, the "Average" method was applied, using the average time based on task type. The machine learning algorithms applied for predicting QC work speed were Linear Regression (LR), Random Forest (RF), Support Vector Regression (SVR), and Multi-Layer Perceptron (MLP).

The experiments were conducted based on actual port operation schedules to measure prediction errors. These results highlight the impact of using ML models on improving accuracy in predicting QC work speeds and the resulting operational plans.

**Table 2. Prediction error of QC work speed prediction**

| Methods | QC Work Speed | | Task work time | | Makespan | |
|---|---|---|---|---|---|---|
| | MAE | MAPE | MAE | MAPE | MAE | MAPE |
| Constant | 47.41 | 0.37 | 1084 | 0.21 | 8980 | 0.12 |
| Average | 45.19 | 0.39 | 480 | 0.1 | 2761 | 0.04 |
| LR | 44.17 | 0.38 | **419** | **0.09** | 2415 | 0.04 |
| RF | 44.32 | 0.38 | **419** | **0.09** | **2388** | **0.03** |
| SVR | **42.7** | **0.33** | 734 | 0.14 | 8082 | 0.11 |
| MLP | 45.69 | 0.4 | 508 | 0.1 | 2891 | 0.04 |

According to Table 2, the prediction of the time required for a QC to handle a single task showed that SVR had the highest performance, while the constant setting resulted in the largest error. However, for the prediction of task work time and makespan, SVR exhibited larger errors, indicating overfitting to specific working conditions. In contrast, models such as Average, Linear Regression, and Random Forest demonstrated that overfitting was not an issue.

Ultimately, Random Forest showed the highest predictive accuracy for the expected vessel operation time. In comparison, the constant setting (90 seconds for discharge and 150 seconds for loading) resulted in significantly larger prediction errors for task work time and makespan as the number of containers increased, compared to data-driven methods.

When there is a significant difference between the estimated and actual operation times during work plan generation, the plan may need to be revised, which creates operational inefficiencies. Therefore, the results of this experiment emphasize the importance of leveraging collected data when generating work plans to improve accuracy and reduce the need for revisions.

### 5.2. Fitness Score Prediction

To apply the Surrogate Moel proposed in this study for solving QCSP, it is necessary to have an ML model trained to predict the fitness function. This section summarizes the experimental results aimed at identifying the most suitable ML model for QCSP. The ML models used for learning the fitness function include LR, Generalized Additive Model (GAM), RF, SVR, and MLP.

For the experiment, random QCSP instances were generated based on the bay plans of ships included in the training set, and the models were tested on the test set. A total of 170 ship bay plans were collected, with 80% of the data used for the training set and the remaining 20% for the test set. This experiment aimed to evaluate the performance of these models in predicting the fitness function, which is crucial for improving the efficiency of the Surrogate Model in solving QCSP.

Table 3. Fitness score predictions error

| Methods | Fitness function prediction | |
| --- | --- | --- |
| | MAE | MAPE |
| LR | 3.77 | 0.14 |
| GAM | 3.76 | 0.14 |
| RF | 2.61 | 0.12 |
| **SVR** | **1.44** | **0.08** |
| MLP | 8.66 | 0.48 |

Table 3 presents the learning outcomes of various ML models applied to randomly generated QC work plans (chromosomes). The results show that Support Vector Regressor (SVR) demonstrated the highest performance. It was observed that QC work plans exhibit a nonlinear relationship with individual QC task allocations, and SVR effectively captured this nonlinear pattern, resulting in superior performance.

In contrast, the Multi-Layer Perceptron (MLP) faced issues with convergence during training, which may be attributed to overfitting. This overfitting likely occurred because a wide variety of similar chromosomes were used for training on a single bay plan, causing the model to fail in generalizing effectively.

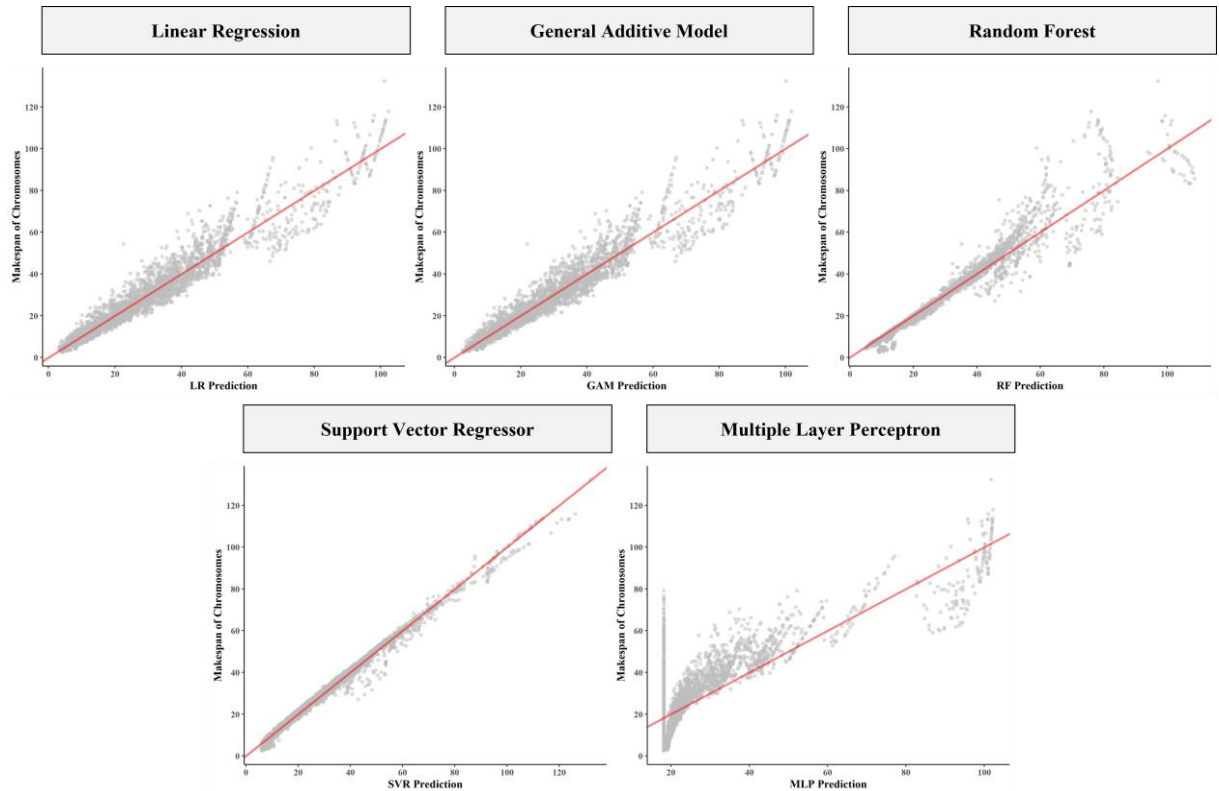

Figure 7. Prediction results of fitness score

Figure 7 illustrates the results of predicting the makespan for the generated QC work plans (chromosomes) using various ML models. The x-axis represents the predicted values, while the y-axis denotes the actual makespan. The results show that SVR provides the most accurate predictions. Based on this, the Surrogate Model for solving QCSP utilizes the trained SVR model to predict the fitness function, ensuring optimal performance in the chromosome evaluation process.

## 5.3. Surrogate Model

Finally, we summarize the experimental results of the Surrogate Model for solving QCSP, focusing on the computational time and best score comparison.

・ The first comparison group employed a GA method designed to speed up the search process using a two-point partially matched crossover operator and a tournament selection strategy to choose the best chromosomes.
・ The second comparison group used a different approach to explore optimal solutions, applying roulette wheel selection, order crossover, and swap mutation methods.

The experiments were conducted by varying the number of chromosomes in the initial population. While increasing the population size improves the chances of finding better chromosomes, it also leads to a significant increase in computation time due to the need for extensive evaluations. The advantage of the Surrogate Model is that it uses the trained ML model to rank the chromosomes and start the search, which makes it less dependent on the initial population size. This reduces the impact of the initial population on search performance, allowing for efficient exploration without significantly increasing computational cost.

The Surrogate Model was set up with the following configuration

**Table 4. Parameters for Surrogate Model**

| Descriptions | Notations | Value |
| --- | --- | --- |
| Maximum generation | $G$ | 10 |
| Number of parents choosing | $np$ | 5 |
| Maximum evolution | $e$ | 10 |

As summarized in Table 4, the experimental parameters for the Surrogate Model were set as follows: the maximum generation ($G$) was 10, the number of parent chromosomes selected ($np$) was 5, and the maximum number of evolutions ($e$) was 10. The initial population sizes were set at 100, 300, 500, and 700.

The experiment was conducted on a test set consisting of 34 bay plans, and the results are summarized in Table 5. These results compare the performance of the Surrogate Model across different initial population sizes, highlighting the trade-offs between computation time and the quality of the best solution (best score). The findings confirm that the Surrogate Model's efficiency is relatively unaffected by the size of the initial population, due to its ability to rank chromosomes using the ML model before starting the search. This allows for effective exploration with reduced computation time, even with smaller initial populations.

According to Table 5, the experimental results are organized based on the initial population size and the total container workload of the bay plan (VAN), showing the average computing time and the average score of the best chromosome. The results reveal that when the initial population size is 100, the computational speed of the other GA algorithms is faster than that of the Surrogate Model. However, when comparing the scores obtained within that time frame, the chromosomes identified by the Surrogate Model achieve better scores.

As the initial population size increases, the Surrogate Model demonstrates a significant advantage in terms of computational efficiency. Unlike other GA methods, the computational time for the Surrogate Model does not increase drastically with larger population sizes, showing that it maintains its efficiency while consistently delivering high-quality solutions. This efficiency is due to the ML model's ability to rank chromosomes, allowing the Surrogate Model to focus on evaluating the most promising solutions first.

Table 5. Experiment results of QCSP

| Initial population | Total workload (VAN) | GA [9] Time | GA [9] Score | GA [22] Time | GA [22] Score | Surrogate Model Time | Surrogate Model Score |
|---|---|---|---|---|---|---|---|
| 100 | ~ 299 | 136.61 | 0.23 | **59.81** | 0.27 | 81.72 | **0.07** |
| | 300 ~ 599 | 271.54 | 2.27 | **115.69** | 2.09 | 179.62 | **1.64** |
| | 600 ~ 999 | 427.82 | 9.53 | **184.46** | 9.56 | 442.6 | **9.14** |
| | 1000 ~ 1499 | 713.33 | 23.18 | **302.5** | 24.6 | 1162.85 | **19.22** |
| | 1500 ~ | 770.71 | 20.92 | 312.08 | 20.28 | 553.01 | **16.35** |
| 300 | ~ 299 | 408.42 | 0.19 | 175.28 | 0.23 | **86.62** | **0.11** |
| | 300 ~ 599 | 801.55 | 1.94 | 342.26 | 1.78 | **201.02** | **1.44** |
| | 600 ~ 999 | 1336.21 | 10.3 | 554.67 | 9.12 | **239.31** | **8.98** |
| | 1000 ~ 1499 | 2123.6 | 21.31 | **879.89** | 20.94 | 1070.55 | **18.15** |
| | 1500 ~ | 2266.89 | **15.41** | 940.26 | 20.68 | 351.34 | 16.69 |
| 500 | ~ 299 | 692.02 | 0.16 | 291.96 | 0.13 | **94.49** | **0.1** |
| | 300 ~ 599 | 1346.19 | **1.37** | 568.53 | 1.79 | **187.82** | 1.41 |
| | 600 ~ 999 | 2158.41 | 9.38 | 919.33 | 9.52 | **471.87** | **8.69** |
| | 1000 ~ 1499 | 3461.46 | 21.09 | 1468.41 | 20.48 | **1199.84** | **17.79** |
| | 1500 ~ | 3792.61 | 17.91 | 1602.01 | 18.61 | **425.09** | **15.94** |
| 700 | ~ 299 | 901.05 | 0.13 | 413.21 | **0.07** | 113.12 | 0.12 |
| | 300 ~ 599 | 1805.73 | 1.55 | 799.85 | 1.63 | **213.79** | **1.4** |
| | 600 ~ 999 | 2881.71 | **8.74** | 1279.08 | 9.97 | **306.1** | 8.78 |
| | 1000 ~ 1499 | 4638.46 | 19.89 | 2093.83 | 19.74 | **1190.84** | **17.56** |
| | 1500 ~ | 4995.2 | 17.2 | 2263.25 | 18.46 | **507.11** | **15.94** |

According to Figure 8, as the initial population size increases, other GA methods require more time to evaluate a larger number of chromosomes. In contrast, the Surrogate Model maintains a consistent computational pattern because it uses the pre-trained ML model to rank the chromosomes in advance based on their predicted fitness. This shows a significant advantage in reducing search time, demonstrating that applying an ML model in GA can dramatically decrease the computation time.

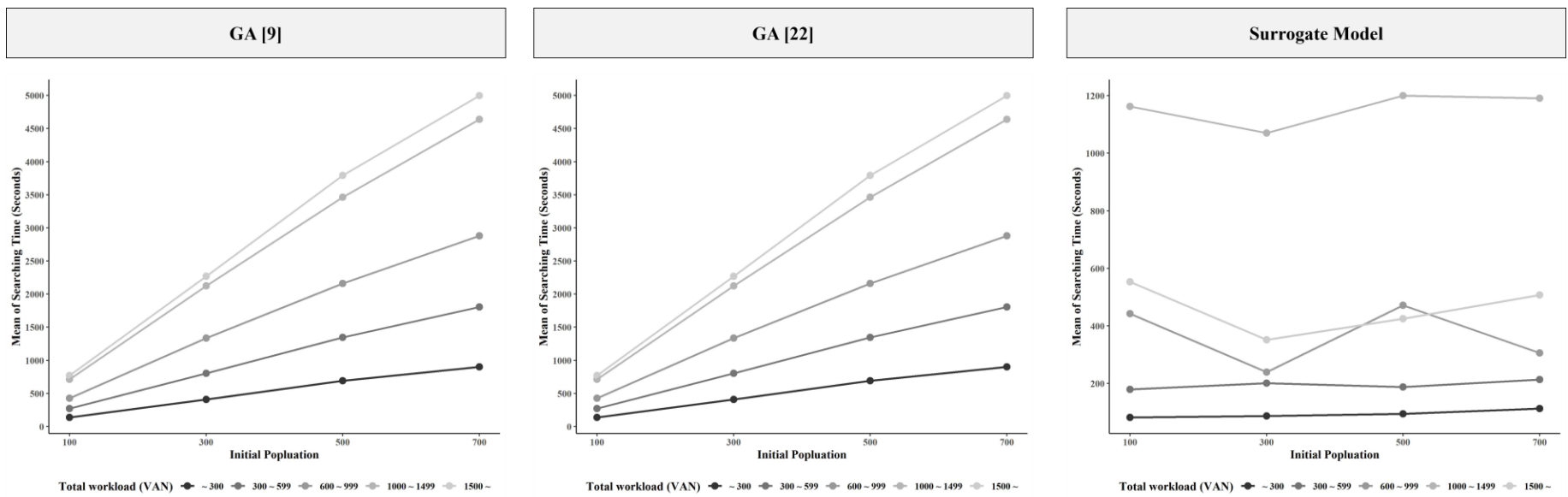

**Figure 8. Computational time according to the size of initial populations**

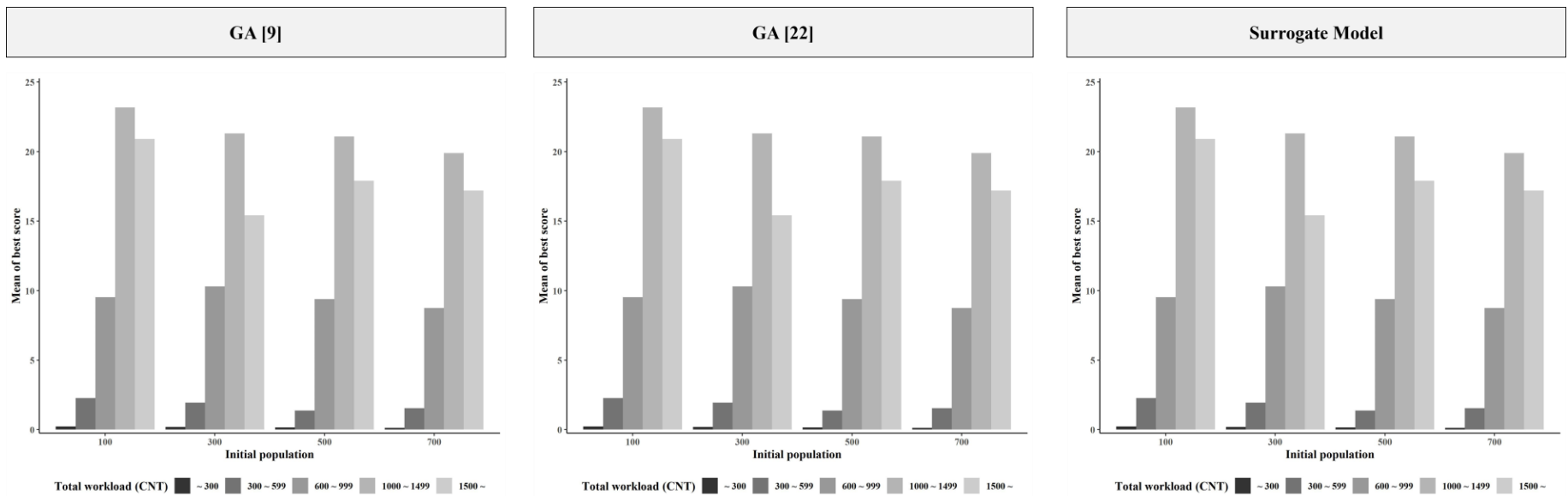

**Figure 9. Fitness score according to the size of initial populations**

Figure 9 summarizes the best scores achieved by each GA methodology based on the initial population size and the number of containers processed. As expected in GAs, searching through a larger population generally leads to finding chromosomes with better fitness scores. However, as noted in Figure 8, other GA methods show a proportional increase in computational time as the initial population grows, which is a significant drawback.

In contrast, the Surrogate Model offers the advantage of being able to set a larger initial population without a corresponding increase in computation time. This is due to the pre-trained ML model, which ranks the chromosomes in advance and focuses the search on the most promising candidates, thereby improving search efficiency. To summarize the experimental results, when solving QCSP and other similar NP-hard problems, combining heuristic algorithms with an ML model, as demonstrated by the Surrogate Model, enables more efficient global search. This approach reduces computational costs while maintaining or even improving the quality of the solutions found, proving to be a highly effective method for tackling complex optimization problems.

## 6. Conclusion

The Quay Crane Scheduling Problem (QCSP) is a critical aspect of planning operations in container ports, and numerous studies have been conducted to address this challenge. Previous research has proposed various methods to solve large-scale problems, and this study follows that trajectory by utilizing bay plan data collected from Busan Port Terminal (South Korea) to develop an algorithm applicable in real-world situations.

This study introduces a method for creating QC work plans by predicting the working speed of QCs, leading to more accurate planning. The working speed of a QC varies depending on the attributes of the container being handled, and by incorporating this into the planning process, we demonstrated that the resulting work plans are more precise. Furthermore, to improve the search speed and performance of GAs, commonly used to solve NP-hard problems, we introduced the Surrogate Model, which combines GA with an ML model. The Surrogate Model leverages an ML model trained on the fitness function, allowing for faster searches and the generation of higher-quality chromosomes compared to previous GA approaches applied to QCSP. This finding suggests that combining heuristic algorithms, widely used for solving NP-hard problems, with ML models, often employed in prediction tasks, can lead to the development of more efficient search algorithms.

However, a limitation discovered in this study is that the Surrogate Model 's performance depends heavily on the accuracy of the ML model's fitness function predictions. If the ML model mispredicts and ranks chromosomes incorrectly, it can slow down the search process compared to traditional GA methods. Therefore, careful attention must be paid to defining the input features for learning the fitness function when combining ML models with GAs.

This study did not incorporate constraints related to vessel stability, integrated scheduling with yard trucks, and other operational factors in QCSP. Future research will focus on integrating these additional constraints to create more comprehensive planning solutions.

**References**

[1]. Ambrosino, D., Sciomachen, A., & Tanfani, E. (2004). Stowing a containership: The master bay plan problem. *Transportation Research Part A: Policy and Practice, 38*(2), 81-99. https://doi.org/10.1016/j.tra.2003.09.002

[2]. Lee, E., Park, K., Kim, D., Bae, H., & Hong, C. (2021). Prediction of the quay crane's handling time with external handling factors. *ICIC Express Letters, Part B: Applications, 12*(4), 351-358. https://doi.org/10.24507/icicelb.12.04.351

[3]. Bierwirth, C., & Meisel, F. (2015). A follow-up survey of berth allocation and quay crane scheduling problems in container terminals. *European Journal of Operational Research, 244*(3), 675-689. https://doi.org/10.1016/j.ejor.2014.12.030

[4]. Daganzo, C. (1989). The crane scheduling problem. *Transportation Research Part B: Methodological, 23*(3), 159-175. https://doi.org/10.1016/0191-2615(89)90001-5

[5]. Kim, K., & Park, Y.-M. (2004). A crane scheduling method for port container terminals. *European Journal of Operational Research, 156*(3), 752-768. https://doi.org/10.1016/S0377-2217(03)00133-4

[6]. Chung, S.-H., & Choy, K. L. (2012). A modified genetic algorithm for quay crane scheduling operations. *Expert Systems with Applications, 39*(4), 4213-4221. https://doi.org/10.1016/j.eswa.2011.09.113

[7]. Liu, M., Liang, B., Zheng, F., Chu, C., & Chu, F. (2018). Quay crane scheduling problem with the consideration of maintenance. In *2018 IEEE International Conference on Networking, Sensing and Control (ICNSC)* (pp. 1-6). IEEE. https://doi.org/10.1109/ICNSC.2018.8361346

[8]. Al-Dhaheri, N., & Diabat, A. (2015). The quay crane scheduling problem. *Journal of Manufacturing Systems, 36*, 87-94. https://doi.org/10.1016/j.jmsy.2015.02.010

[9]. Zhang, Z., Liu, M., Lee, C.-Y., & Wang, J. (2018). The quay crane scheduling problem with stability constraints. *IEEE Transactions on Automation Science and Engineering, 15*(4), 1565-1578. https://doi.org/10.1109/TASE.2018.2795254

[10]. Wu, L., & Ma, W. (2017). Quay crane scheduling with draft and trim constraints. *Transportation Research Part E: Logistics and Transportation Review, 97*, 98-112. https://doi.org/10.1016/j.tre.2016.10.011

[11]. Hu, H., Chen, X., Zhen, L., Ma, C., & Zhang, X. (2019). The joint quay crane scheduling and

block allocation problem in container terminals. *IMA Journal of Management Mathematics, 30*(1), 51-75. https://doi.org/10.1093/imaman/dpy013

[12]. Behjat, N., & Nahavandi, N. (2020). Quay cranes and yard trucks scheduling problem at container terminals. *Iranian Journal of Engineering, 33*(9), 1751-1758. https://doi.org/10.5829/ije.2020.33.09c.08

[13]. Xingchi, W., Caimao, T., & Junliang, H. (2019). Integrated optimization of berth allocation and quay crane assignment under uncertainty. In *2019 IEEE International Conference on Advanced Computing and Machine Learning (ICACMVE)* (pp. 11-16). IEEE. https://doi.org/10.1109/ICACMVE.2019.00011

[14]. Abou Kasm, O., & Diabat, A. (2020). Next-generation quay crane scheduling. *Transportation Research Part C: Emerging Technologies, 114*, 694-715. https://doi.org/10.1016/j.trc.2020.02.015

[15]. Boysen, N., Briskorn, D., & Meisel, F. (2016). A generalized classification scheme for crane scheduling with interference. *European Journal of Operational Research, 254*(1), 127-138. https://doi.org/10.1016/j.ejor.2016.08.041

[16]. Sun, D., Tang, L., Baldacci, R., & Lim, A. (2020). An exact algorithm for the unidirectional quay crane scheduling problem with vessel stability. *European Journal of Operational Research, 291*(3), 861-874. https://doi.org/10.1016/j.ejor.2020.09.033

[17]. Phan-Thi, M.-H., Ryu, K., & Kim, K. (2013). Comparing cycle times of advanced quay cranes in container terminals. *Industrial Engineering and Management Systems, 12*(4), 359-369. https://doi.org/10.7232/iems.2013.12.4.359

[18]. Al-Dhaheri, N., Jebali, A., & Diabat, A. (2016). A simulation-based genetic algorithm approach for quay crane scheduling under uncertainty. *Simulation Modelling Practice and Theory, 66*, 122-138. https://doi.org/10.1016/j.simpat.2016.01.009

[19]. UNCTAD. (2019). *Review of Maritime Transport 2019*. United Nations Conference on Trade and Development. https://unctad.org/webflyer/review-maritime-transport-2019

[20]. He, J. (2016). Berth allocation and quay crane assignment in a container terminal for the trade-off between time-saving and energy-saving. *Advanced Engineering Informatics, 30*(3), 390-405. https://doi.org/10.1016/j.aei.2016.04.006

[21]. Wang, T., Du, Y., Fang, D., & Li, Z.-C. (2020). Berth allocation and quay crane assignment for the trade-off between service efficiency and operating cost considering carbon emission taxation. *Transportation Science, 54*(4), 1046-1065. https://doi.org/10.1287/trsc.2019.0946

[22]. Kenan, N., Jebali, A., & Diabat, A. (2021). The integrated quay crane assignment and scheduling problems with carbon emissions considerations. *Computers & Industrial Engineering, 157*, 107734. https://doi.org/10.1016/j.cie.2021.107734

[23]. He, J., Wang, Y., Tan, C., & Yu, H. (2021). Modeling berth allocation and quay crane assignment considering QC driver cost and operating efficiency. *Advanced Engineering Informatics, 47*, 101252. https://doi.org/10.1016/j.aei.2021.101252

[24]. Lee, D.-H., Wang, H., & Miao, L. (2008). Quay crane scheduling with non-interference constraints in port container terminals. *Transportation Research Part E: Logistics and Transportation Review, 44*(1), 124-135. https://doi.org/10.1016/j.tre.2006.08.001

[25]. Rodrigues, F., & Agra, A. (2021). Berth allocation and quay crane assignment/scheduling problem under uncertainty: A survey. *European Journal of Operational Research*. https://doi.org/10.1016/j.ejor.2021.01.047

[26]. Fu, Y.-M., Diabat, A., & Tsai, I.-T. (2014). A multi-vessel quay crane assignment and scheduling problem: Formulation and heuristic solution approach. *Expert Systems with Applications, 41*(15), 6959-6965. https://doi.org/10.1016/j.eswa.2014.05.002

[27]. Legato, P., & Mazza, R. M. (2001). Berth planning and resources optimisation at a container terminal via discrete event simulation. *European Journal of Operational Research, 133*(3), 537-547. https://doi.org/10.1016/S0377-2217(00)00200-9

[28]. Cahyono, R. T., Flonk, E. J., & Jayawardhana, B. (2020). Discrete-event systems modeling and the model predictive allocation algorithm for integrated berth and quay crane allocation. *IEEE Transactions on Intelligent Transportation Systems, 21*(3), 1321-1331. https://doi.org/10.1109/TITS.2019.2910283

[29]. Chen, J. H., Lee, D.-H., & Cao, J. X. (2012). A combinatorial benders' cuts algorithm for the quayside operation problem at container terminals. *Transportation Research Part E: Logistics and Transportation Review, 48*(2), 266-275. https://doi.org/10.1016/j.tre.2011.06.004

[30]. Park, K., Sim, S., & Bae, H. (2021). Vessel estimated time of arrival prediction system based on a path-finding algorithm. *Maritime Transport Research, 2*, 100034. https://doi.org/10.1016/j.martra.2021.100034

[31]. Chargui, K., Zouadi, T., El Fallahi, A., Reghioui, M., & Aouam, T. (2021). A quay crane productivity predictive model for building accurate quay crane schedules. *Supply Chain Forum: An International Journal, 22*(2), 136-156. https://doi.org/10.1080/16258312.2020.1831889

[32]. Chahar, V., Katoch, S., & Chauhan, S. (2021). A review on genetic algorithm: Past, present, and future. *Multimedia Tools and Applications, 80*(14), 19605-19637. https://doi.org/10.1007/s11042-020-10139-6

[33]. Matos Dias, J., Rocha, H., Ferreira, B., & Lopes, M. D. C. (2014). A genetic algorithm with neural network fitness function evaluation for IMRT beam angle optimization. *Central European Journal of Operations Research, 22*(3), 507-528. https://doi.org/10.1007/s10100-013-0289-4

[34]. Kwon, Y., Kang, S., Choi, Y.-S., & Kim, I. (2021). Evolutionary design of molecules based on

deep learning and a genetic algorithm. *Scientific Reports, 11*(1), 17442. https://doi.org/10.1038/s41598-021-96812-8

[35]. Radaideh, M., & Shirvan, K. (2021). Rule-based reinforcement learning methodology to inform evolutionary algorithms for constrained optimization of engineering applications. *Knowledge-Based Systems, 217*, 106836. https://doi.org/10.1016/j.knosys.2021.106836